\documentclass{article}

\usepackage{microtype}
\usepackage{graphicx}
\usepackage{subfigure}
\usepackage{booktabs} 

\usepackage{hyperref}

\usepackage[accepted]{icml2021}

\icmltitlerunning{Towards Automated Evaluation of Explanations in Graph Neural Networks}

\begin{document}

\twocolumn[
\icmltitle{Towards Automated Evaluation of Explanations in Graph Neural Networks}




\begin{icmlauthorlist}
\icmlauthor{Vanya BK}{iitm}
\icmlauthor{Balaji Ganesan}{irl}
\icmlauthor{Aniket Saxena}{ibm}
\icmlauthor{Devbrat Sharma}{isl}
\icmlauthor{Arvind Agarwal}{irl}
\end{icmlauthorlist}

\icmlaffiliation{iitm}{IIT Madras, India}
\icmlaffiliation{irl}{IBM Research, India}
\icmlaffiliation{ibm}{IBM, India}
\icmlaffiliation{isl}{IBM Data and AI, India}

\icmlcorrespondingauthor{Balaji Ganesan}{bganesa1@in.ibm.com}

\icmlkeywords{Machine Learning, ICML}

\vskip 0.3in
]



\printAffiliationsAndNotice{}  

\begin{abstract}
Explaining Graph Neural Networks predictions to end users of AI applications in easily understandable terms remains an unsolved problem. In particular, we do not have well developed methods for automatically evaluating explanations, in ways that are closer to how users consume those explanations. Based on recent application trends and our own experiences in real world problems, we propose automatic evaluation approaches for GNN Explanations.
\end{abstract}

\section{Introduction}
Explaining neural model predictions to the users of an application adds tremendous value, enhancing trust in the model predictions and increases adoption of AI even in sensitive real world applications.

While explainability techniques in other areas of machine learning  like LIME \cite{ribeiro2016should}, Influence functions \cite{koh2017understanding}, SHAP \cite{lundberg2017unified}, and Anchors \cite{ribeiro2018anchors} have been widely adopted in real world applications, adoption of graph neural networks (GNN) explainability techniques has faced some challenges.

Recent works in GNN explanations include \cite{ying2019gnnexplainer}, \cite{yuan2020xgnn}, \cite{vu2020pgmexplainer}, \cite{yuan2021explainability}, \cite{schlichtkrull2020interpreting}. Many of them tend to produce a subgraph of important features, nodes and edges, as explanation for a GNN model prediction. Let us collectively call these type of explanations as \textit{subgraph explanations}.

\begin{figure}[!htp]
    \centering
    \includegraphics[width=\columnwidth]{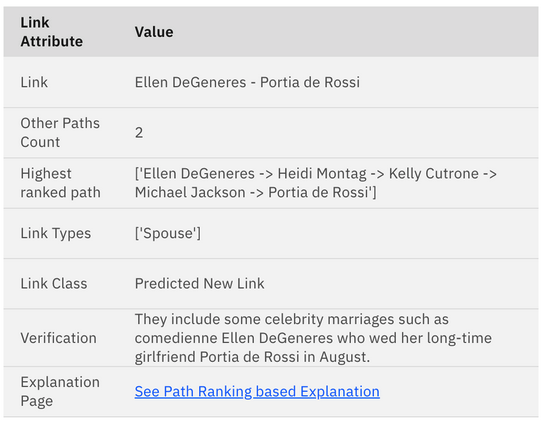}
    \caption{An example GNN explainability implementation that uses features, paths, and verifiable text to substantiate a prediction}
    \label{fig:xlp}
\end{figure}

Our experience from real world applications is that these subgraph explanations are usually hard for end users to understand. Even for model developers it is difficult to determine what is an acceptable explanation for a prediction. In industrial applications, this problem is sought to be solved with user-studies \cite{ganesan2020explainable} and active learning based approaches.

However, designing such user-studies could be cost prohibitive, take lot of time and effort, and still may not lead to desired outcomes. Explainability is subjective and users may not be able to provide feedback unless different explanations are presented to them. Further, different human annotators who annotate explanations or verify the correctness of annotations may have significant knowledge of the domain or the corpus, which makes them overlook missing parts of an explanation. So what may be a reasonable explanation to one user might be indecipherable to another. 

Hence we need a framework to automatically evaluate explanations before they are presented to human evaluators.

In law courts, it is said that the best case from the point of view of prosecutors, is a strong circumstantial case. Circumstantial cases have a number of objectively provable facts. Each of those facts may not be enough, but together they make a strong case. Even more than cases that have witness accounts, which are called testimonial evidence, because witnesses may have motives, or may not be sure of their own testimony. Evaluation of explainability techniques could benefit from this insight.

This leads us to explore different automated techniques which can substantiate elements of an explanation, each of which are easy to generate and verify, like circumstantial evidence, but may not be enough individually as an explanation.

Consider the link prediction task using GNNs, where two nodes in a graph are predicted to be linked. Examples of techniques that can substantiate a predicted link include, highlighting important features and neighbors of the nodes in the form of a subgraph \cite{ying2019gnnexplainer}, a sentence from a news article or Wikipedia page that mentions the predicted link \cite{ganesan2020link}, displaying other paths that exist between the two nodes \cite{ganesan2020explainable}, and generating examples from the training data that are similar to the current prediction.

Accumulating multiple solutions to solve a problem is well known to the machine learning community. Whether it's ensemble models, or resolving differences between multiple annotators, or the data programming paradigm used in Snorkel \cite{ratner2017snorkel}, we have used a combination of methods when working under uncertainty.

This extended abstract is organized as follows. In Section \ref{global_structure}, we discuss one of the reasons why GNN explanations may not be that satisfactory, namely the lack of global structure information. In Section \ref{real_world}, we present a number of approaches we have tried in a real world application and our ideas to take those approaches forward. Finally, in Section \ref{nsai}, we formalize our evaluation framework based on recent work in neuro-symbolic reasoning.

\section{Global structure information}
\label{global_structure}

In graph neural networks, a node is represented by its features and by aggregating information from its corresponding neighbor nodes which lie within a limited neighboring region. This way of aggregating information, using the message passing framework, can be considered as capturing the interactions between different graph structures. For graph-level tasks (like graph classification) and to some extent for node-level tasks (like node classification and link prediction), it is imperative to capture \textit{global structure} to provide better explanations.

Assuming our input graph is L-hop separated, we can use L layers of GNN in our graph model \textit{f($\bullet$)} so as to capture the global structure of the input graph. Therefore, a GNN model can be given as 
\begin{equation}
    E = f(G, X)
\end{equation}
where G is the input graph, X is the adjacency matrix associated with G, and E is the matrix of final node representations containing both features and the overall structure of the input graph G. These final node representations can further be utilized to help providing useful explanations.

Recently, Multi-Layer Perceptrons (MLPs) have extensively been used as an explanation network in masking-based explanation methods \cite{luo2020parameterized,schlichtkrull2020interpreting} to identify the edges which are affecting final predictions the most. Moreover, the final node representations from the original GNN model can be given as an input to MLP to learn a masking function which further assigns weights to every single edge of the input graph to decide which edges are important for a particular prediction
\begin{equation}
    e_{ij} = MLP_{\theta}(E)
\end{equation}
where $e_{ij}$ is the weight associated with the edge between node \textit{i} and \textit{j} computed by MLP.

The reason for using MLP is two-fold: (i) the parameters ($\theta$) of MLP are shared by all edges of the input graphs and thus help in capturing the global view of the graph model; (ii) approximation of derivatives of the learned function can be done by derivatives of MLPs \cite{Goodfellow-et-al-2016}.

However, these methods do not take into account the interactions among different edges and nodes. Also the explanation using these methods, may result in disconnected nodes and edges, thereby become less intelligible to end users. In contrast to aforementioned perturbation-based methods, non-perturbation-based methods do not utilize MLP. Rather, these methods aim to consider different subgraphs of the input graph. The most recent subgraph-based approach is SubgraphX \cite{yuan2021explainability}, which aims to explore different subgraphs and identify only those that mostly impact the final predictions. Although SubgraphX assures to offer connected subgraphs as explanation for every single input, this approach is highly sensitive to actions being considered while looking for a path using Monte-Carlo Tree Search (MCTS) being utilized for SubgraphX method. This is because different actions cause SubgraphX to output different explanations.


Here we observe that further work in this direction may result in subgraph explanations that are more intelligible to end users. However, such methods still need to be evaluated, preferably using automated evaluation techniques, before they are included in user-studies and production deployments (to reduce the cost associated with human annotations). In the next section, we'll discuss some of the techniques we have used in industrial applications to assist human annotators, that can serve this automated evaluation purpose as well.

\begin{figure}[!htp]
    \centering
    \includegraphics[width=\columnwidth]{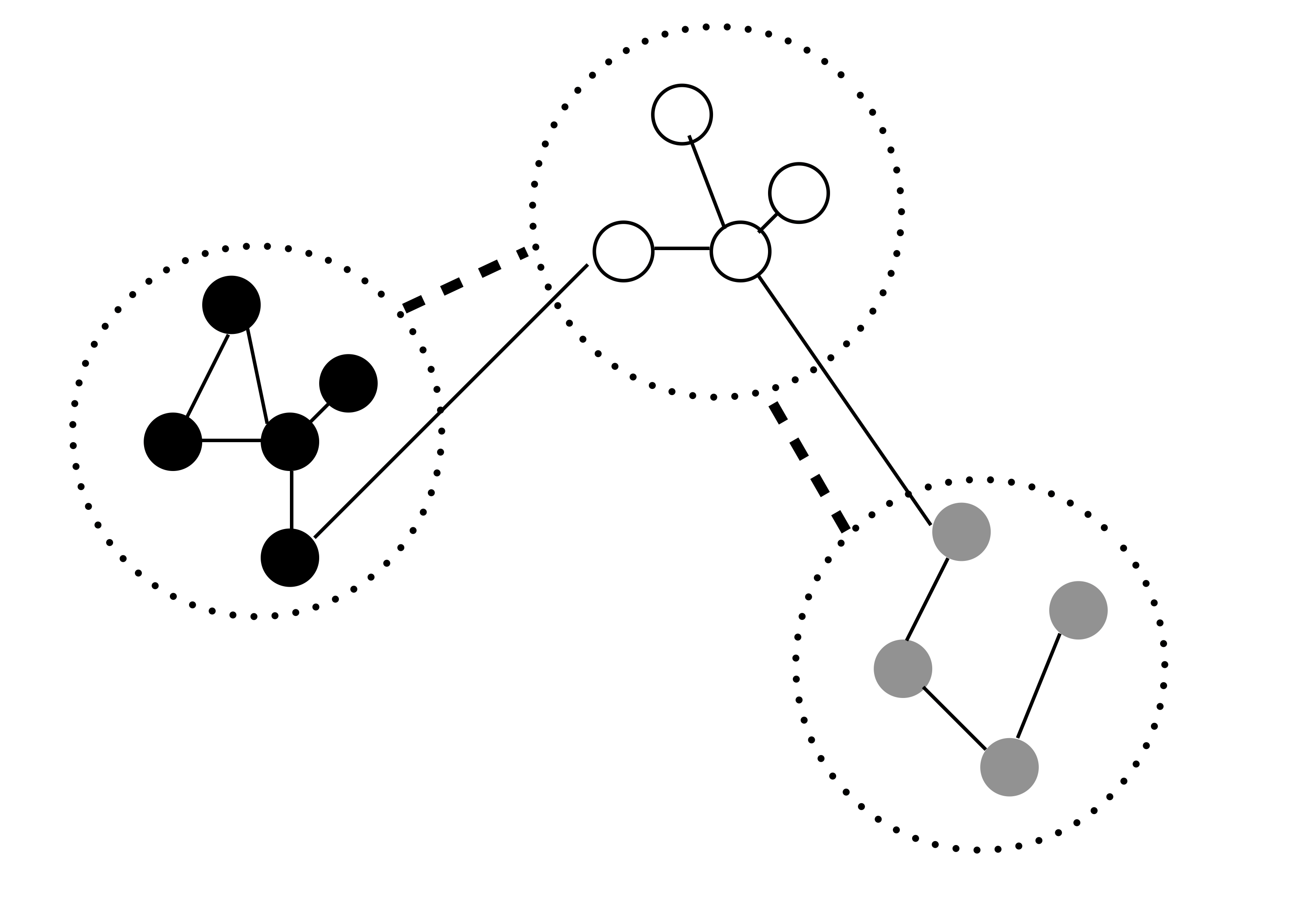}
    \caption{An example of graph clustering based explanation that can substantiate GNN model explanations}
    \label{fig:clustering_label}
\end{figure}

\begin{figure*}[htp]
    \centering
    \includegraphics[width=\linewidth]{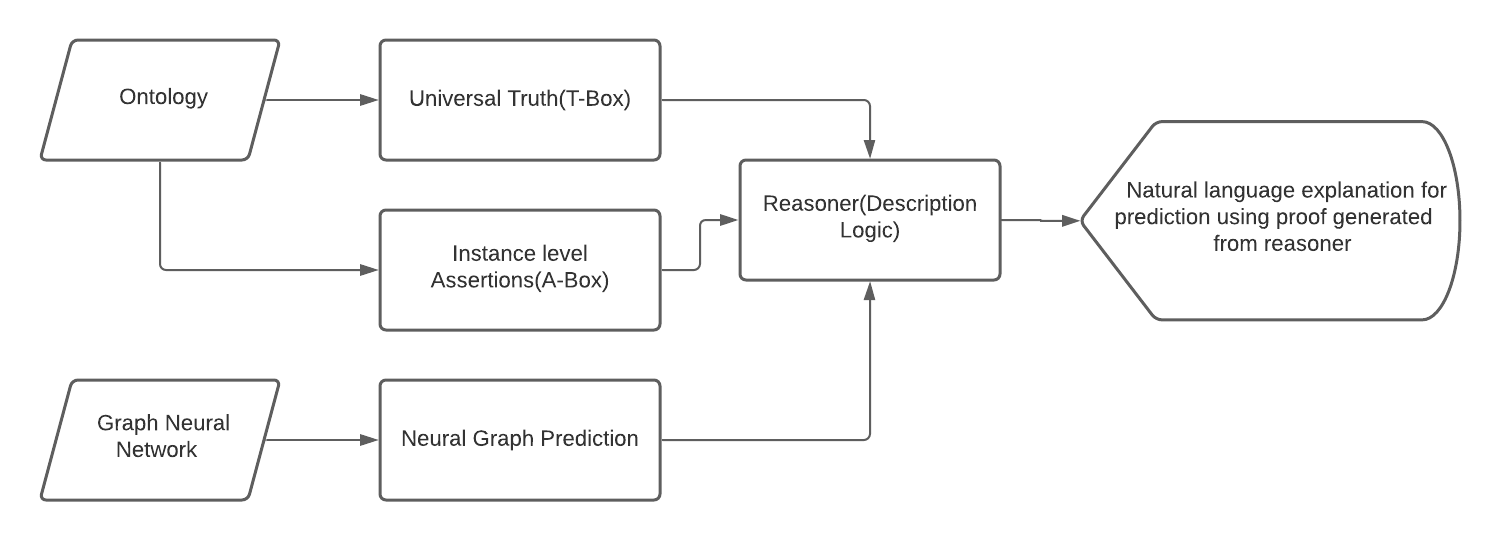}
    \caption{Reasoning mechanism}
    \label{fig:my_label}
\end{figure*}

\section{Clustering as Explanation}
\label{real_world}

In \cite{ganesan2020link} and \cite{ganesan2020explainable}, we had discussed few approaches using information retrieval and graph algorithms, which can be used to substantiate the explanations from model-based GNN explanations.

In Figure \ref{fig:xlp}, we've shown two of these techniques. This was a graph of people that we populated using the triples in the TACRED dataset \cite{zhang2017position}. A link between the nodes \textit{Ellen DeGeneres} and \textit{Portia de Rossi}, could be substantiated by displaying two longer paths that already exist between them. We have also displayed the highest ranked path among them. Next, we indexed and retrieved a text from the same TACRED dataset. We could also have found evidence from Wikipedia or news articles that could substantiate the GNN model prediction. 

An extension of this idea is to use clustering techniques to generate explanations. Consider the problem of entity matching in property graphs (homogeneous attributed graphs). This is determining if two nodes in a graph are infact the same node and hence can be merged. More generally, this can be considered as a link prediction problem with link type \textit{same\_as}. This can also be thought of as a graph similarity problem where the attributes of nodes are nodes themselves like in a knowledge graph.

In industrial applications, clustering techniques have previously been used exploratory analysis to find similarities within the data, draw inferences, find hidden patterns and also to reconstruct possibly underlying data structures.

Different methods have been proposed in the literature cluster nodes in a graph. By treating nodes as rows in relational data stored in an external database and clustering the rows could be one approach. Graph clustering that uses traditional graph algorithms, edge weights etc to cluster nodes. Graph variational auto-encoders \cite{kipf2016variational} have also been used to clusters nodes in the graph. More recently GNN embeddings have been used for clustering \cite{desai2021graph}. Once the clusters are generated using any of these methods, the embeddings can be projected on a plane to visualize the nodes.



Consider the example in Figure \ref{fig:clustering_label}. Let us assume that all edges in this graph have been predicted by a GNN model trained for same\_as relation. One way to understand what could be happening is to cluster these nodes. Now, let us further assume that there are three clusters shown as dotted lines in the figure. The dark lines indicate that nodes with the cluster share some common features, and the dotted lines indicate that each cluster may share some abstract or fewer features with other clusters.

By using perturbations, we might be able drop many features and identify important features. In \cite{vannur2021data}, we had used LIME and SHAP to identify important features by treating nodes as rows in relational data. So let us assume that we're able to label the lines in Figure \ref{fig:clustering_label} with important features responsible for those links.

Now, let's revisit the main argument in this work. We propose that the above clustering technique can be used to substantiate parts of the explanations we might get from subgraph explanations from GNN explainability techniques. If the nodes and edges highlighted in the subgraph explanations have lot of overlap with the important features identified in clustering, we'll have greater confidence in the GNN model explanations.


\section{Reasoning based approaches}
\label{nsai}

In this section, we'll discuss a way in which we could formalize the automatic evaluation process. We propose that we can use reasoners typically used in semantic web and ontologies, to automatically evaluate the explanations from GNN explainability techniques, provided we're able to convert them into axioms.


A neuro-symbolic reasoning based approach to this problem as shown in Figure \ref{fig:my_label}, will involve using description logic to represent the universal truth (axioms) in terms of concepts and relations between them. 

\begin{figure}[htp]
    \centering
    \includegraphics[scale=0.2]{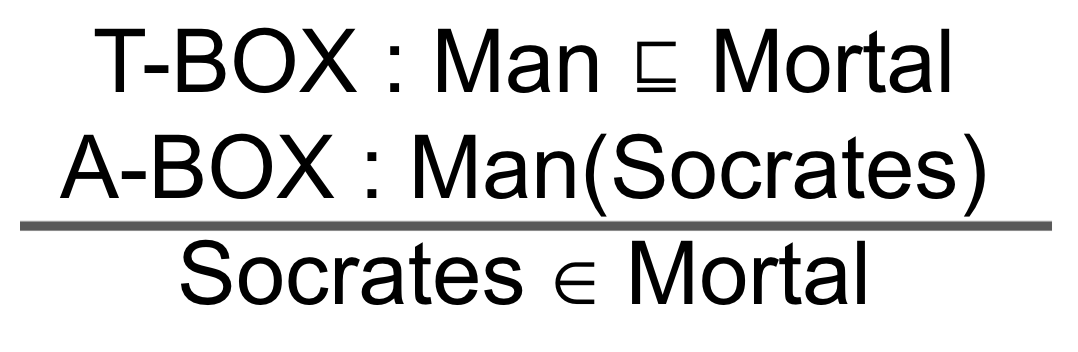}
    \caption{Example of conclusion by reasoner}
    \label{fig:my_example}
\end{figure}

Typically a reasoner (which is complete i.e, can prove all statements that are true given the knowledge base) is fed the axioms (universal truth) along with the instance level axioms, and the goal. Then using several rules of inference such as \textit{modus ponens}, \textit{modus tollens}, rules of simplification, addition etc, the reasoner can generate the proof for the given goal. Optionally this proof in terms of concepts and relations can also be converted as natural language explanations.

Now, let's see how such a reasoner can be used in our use-case. If a reasoner is fed universal axioms (from an ontology or created by a human expert), and axioms generated by us from the neural model predictions, then the reasoner could tell us if those axioms hold true.

As shown in Figure \ref{fig:my_example}, an example rule of modus ponens can be, if "All men are mortal" is the universal truth and "Socrates is a man" is instance level assertion and the neural graph prediction is "Socrates is mortal", then this prediction can be easily concluded using the modus ponens rule of inference which states that if "P implies Q" and P occurs in the knowledge base then Q can be concluded. The complete proof generated would be "All men are mortal. Socrates is a man. Socrates is a mortal", which can be interpreted easily by the end user. The output of the reasoner, could itself be quantified to evaluate the subgraph explanations. In the next section, we'll briefly explain natural language explanations.

\subsection*{Natural Language Explanations}
In description logic, certain concept forming operators such as negation, conjunction, disjunction, universal restriction, and existential restriction are used to build the concepts. Let C represent concept description and r represents relations. Then semantics is defined based on an interpretation I = ($\Delta^I$,$.^I$) where $\Delta^I$ represents the domain i.e, set of individuals, and the interpretation function $.^I$ maps each concept C to a set of individuals $C^I$ belonging to $\Delta^I$ and each relation r belongs to a binary relation $\Delta^I\times\Delta^I$.
TBox refers to axioms on concepts(universal truth) whereas ABox refers to assertions on individuals which can be role assertions, e.g, Parent(Bill,Jill) or concept assertions, e.g, Man(Socrates).\\
There are several tools which convert natural language to formal representation using the interpretation I. This can be used by description logic to reason about the given axioms. One such tool is FRED \cite{GangemiEtAl2017}.


Since the description logic are sound (all the conclusions generated by the reasoner are true given the set of axioms in the knowledge base) and complete (all the conclusions which are true given the knowledge base can be proved using the reasoner), we can use the above set of axioms given by FRED (converted from the natural language explanations) to conclude the prediction from the graph neural network. If the prediction can be concluded then we can say that the natural language explanations are valid. Hence, this can be used as a means to evaluate the natural language explanation.

Finally, we note that the natural language explanations generated above could be used in conjunction with the information retrieval approach in \cite{ganesan2020link}. If we have text from the corpus that is semantically similar to the natural language explanations, we have additional data points to substantiate the explanations. Alternatively, such \textit{verification text} from the corpus can help generate the universal axioms needed by the reasoners.


\section*{Conclusion}
In this extended abstract, we have drawn attention to the specific problem of evaluating GNN explanations. We proposed automated methods that could substantiate the explanations from GNN explainability techniques. We argued that the lack of global structure information might be making explanations less understandable, then discussed clustering based solutions and finally presented a formal approach to evaluate explanations.

\clearpage
\section*{Acknowledgements}

The first author's participation in this work has been made possible by IBM's Global Research Mentorship program. We thank Abhishek Seth, Somashekar Naganna, Jaydeep Sen, and Akshay Parekh for the discussions.


\bibliography{icml2021}
\bibliographystyle{icml2021}

\end{document}